\documentclass{article}
\usepackage[utf8]{inputenc}
\usepackage[a4paper,top=3.2cm,bottom=3.5cm,left=3.8cm,right=3.8cm,marginparwidth=1.75cm]{geometry}
\usepackage{caption}
\usepackage{subcaption}
\usepackage{graphicx}
\usepackage{amsmath,amsthm,amssymb}
\usepackage{dsfont}
\usepackage{color}
\usepackage{listings}
\usepackage{tabularx}
\usepackage{mathtools}
\usepackage{xspace}
\usepackage{comment}
\usepackage[affil-it]{authblk}

\newsavebox\affbox

\newtheorem{theorem}{Theorem}[section]

\newtheorem{definition}{Definition}

\DeclareMathOperator*{\rand}{rand}

\begin{document}

\author[1]{Luc Brun}
\author[2]{Benoit Gaüzère}
\author[1]{Sébastien Bougleux}
\author[3]{Florian Yger}
\affil[1]{Normandie Univ, CNRS, ENSICAEN, UNICAEN, GREYC, France} 
\affil[2]{Normandie Univ, UNIROUEN, UNIHAVRE, INSA Rouen, LITIS, Rouen, France}
\affil[3]{PSL Université Paris-Dauphine, LAMSADE, Paris, France}

\title{A new Sinkhorn algorithm with Deletion and  Insertion operations}
%\author{Luc Brun 
 %   \thanks{Electronic address: \texttt{luc.brun@ensicaen.fr}; Corresponding author}
    %\affil{Normandie Univ., ENSICAEN, UNICAEN, CNRS.}
%}
%\author{Benoit Gaüzère and Sébastien Bougleux and Florian Yger }

\date{October 2021}

\maketitle

\section{Introduction}
\label{sec:intro}

\begin{figure}
    \centering
    \begin{subfigure}[m]{0.45\textwidth}%\subfigure[b][$\epsilon$-assignment function]{
    \includegraphics[width=\textwidth]{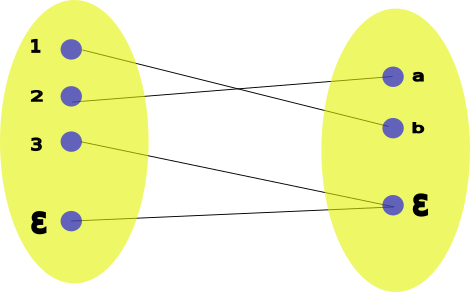}
    \end{subfigure}
    %\subfigure[b][$\epsilon$-assignment matrix]{
    \begin{subfigure}[b]{0.45\textwidth}
    $
    X=\left(\begin{array}{lccc}
    &a&b&\epsilon\\
    \\
    1&0&1&0\\
    2&1&0&0\\
    3&0&0&1\\
    \epsilon&0&0&1\\
    \end{array}
    \right)
    $%}
    \end{subfigure}
    \caption{(a) An example of $\epsilon$-assignment function. 1 is mapped onto b, 2 onto a, 3 is deleted. (b) its associated $\epsilon$ assignment matrix}
    \label{fig:eps_assign}
\end{figure}

This report is devoted to the continuous estimation of an $\epsilon$-assignment(Definition~\ref{def:eps_assign}). Roughly speaking,  an $\epsilon$-assignment between two sets $V_1$ and $V_2$ may be understood as a bijective mapping between a sub part of $V_1$ and a sub part of $V_2$. The remaining elements of $V_1$ (not included in this mapping) are mapped onto an $\epsilon$ pseudo element of $V_2$. We say that such elements are deleted. Conversely, the remaining elements of $V_2$ correspond to the image of the $\epsilon$ pseudo element of $V_1$ (Figure~\ref{fig:eps_assign}). We say that these elements are inserted. 

Let us note that if $V_1$ and $V_2$ have the same size, the bijective mapping induced by an $\epsilon$-assignment may involve all elements of $V_1$, each element being  mapped onto a single element of $V_2$. In this sense, an $\epsilon$ -assignment is more general than a bijective mapping. Moreover, the main advantage of an $\epsilon$-assignment is that it provides us the freedom to not map any element which is then assigned to the $\epsilon$ element of $V_2$ or belong to the image of the $\epsilon$ element of $V_1$. This last property allows us to reject some mappings if for example, these mappings are associated to a large cost. 

An $\epsilon$-assignment function may be associated to an $\epsilon$-assignment matrix (Figure~\ref{fig:eps_assign}(b)) just like any bijective mapping is associated to a permutation matrix. Given two sets, $V_1$ and $V_2$ of respective sizes $n$ and $m$, an $\epsilon$-assignment matrix is encoded by a $(n+1)\times(m+1)$ matrix, where $n+1$ and $m+1$ play respectively the roles of the $\epsilon$ element of $V_1$ and the one of $V_2$. The last column of index $m+1$ of such a matrix encodes the deletions while the last line encodes the insertions. By construction, there is a single $1$ in each of the first $n$ rows and $m$ columns, the remaining elements being set to $0$.

Given $V_1$ and $V_2$, one can define a $(n+1)\times(m+1)$ cost matrix encoding the cost of the mapping of any element of $V_1$ onto an element of $V_2$ as well as the cost of deleting each  element of $V_1$ and inserting each element of $V_2$. Finding an $\epsilon$-assignment minimizing the sum of mappings, deletions and insertions costs is a direct extension of the Linear Sum Assignment Problem (LSAP) called the Linear Sum Assignment Problem with Edition~\cite{TR-bougleux2016} (LSAPE). Given an $\epsilon$-assignment matrix $X$ and a cost matrix $C$, this cost may be formulated as:
\[
\min_{X}\sum_{i=1}^{n+1}\sum_{j=1}^{m+1}c_{i,j}x_{i,j}
\]
where $X$ is taken over all $\epsilon$-assignment matrices. 

We define in previous works~\cite{CI-Bougleux2017,TR-bougleux2016}, an adaptation of the Hungarian algorithm which allows to find an optimal solution to the above problem in $\mathcal{O}(\min(n,m)^2\max(n,m))$. However, while providing an optimal solution, this algorithm does not readily allow the computation of the gradient of the associated operation. This last drawback, does not allow to easily insert such an algorithm into a deep learning pipeline.  On the other hand, the Sinkhorn algorithm~\cite{RI-sinkhorn67}, is based on a continuous relaxation of the problem where permutation matrices are replaced by bi-stochastic matrices with an entropic regularization. This algorithm is the workhorse of computational optimal transport~\cite{peyre2019computational} and is based on iterative matrix multiplications hereby allowing the backpropagation of the gradient~\cite{pmlr-v84-genevay18a}. The aim of this technical report is to transpose the results of the Sinkhorn algorithm to $\epsilon$ assignment matrices. Just like the Sinkhorn algorithm which does not provide a permutation matrix but rather a bi-stochastic matrix, our algorithm will provide an $\epsilon$ bi-stochastic matrix (Definition~\ref{def:bistochastic}). This last point may be of advantage within the Neural Network framework where the hard decisions corresponding to $\epsilon$-assignment matrices may not allow a proper propagation of the gradient. 

More formally, given a similarity matrix $S$ (which may be easily deduced from a cost matrix), we aim at finding two diagonal matrices $D_1$ and $D_2$ such that $D_1SD_2$ is a $\epsilon$ bi-stochastic matrix.  Section~\ref{sec:defNot} provides the main definitions and notations used in the remaining part of this report. The existence and uniqueness of a solution is demonstrated in Section~\ref{sec:ExistUniq} while Section~\ref{sec:construct} provides a constructive algorithm which convergence is demonstrated. Let us note that while Section~\ref{sec:ExistUniq} is a simple adaption of the original proof~\cite{RI-sinkhorn67}, Section~\ref{sec:construct} is significantly different from~\cite{RI-sinkhorn67} since the arguments used for bi-stochastic matrices in the original proof do not hold for $\epsilon$ bi-stochastic matrices. 

\section{Definitions and notations}
\label{sec:defNot}
\begin{definition}[$\epsilon$-assignment]
  \label{def:eps_assign}
  ~\\
  Let $n$ and $m$ be two strictly positive integers. An
  $\epsilon$-assignment is a mapping
  $\varphi: \{1,\dots,n+1\}\rightarrow \mathcal{P}(\{1,\dots,m+1\})$ satisfying the
  following constraints:
  \[
    \left\{
      \begin{array}{ll}
        \forall i\in\{1,\dots,n\},& |\varphi(i)|=1\\
        \forall j\in\{1,\dots,m\},& |\varphi^{-1}(j)|=1\\
        m+1\in \varphi(n+1)\\
      \end{array}
      \right.
    \]
    where $\mathcal{P}(\{1,\dots,m+1\})$ is the power set of $\{1,\dots,m+1\}$. 
  \end{definition}
  Each element of $i\in \{1,\dots,n\}$ is thus mapped onto a set composed of a single element of $\{1,\dots,m+1\}$ $(|\varphi(i)|=1)$ and in the same way the set of antecedents of each $j\in \{1,\dots,m\}$ is reduced to one element $(|\varphi^{-1}(j)|=1)$. Hence the only element of $\{1,\dots,n+1\}$ which can be mapped onto a set composed of several elements  is $n+1$. In the same way, $m+1$ is the only element which may have several antecedents. The constraint $m+1\in \varphi(n+1)$ ensures that $n+1$ is mapped to at least one element and that $m+1$ has at least an antecedent.  
  
  In the example of Figure~\ref{fig:eps_assign} we have $n=3$ and $m=2$. Elements  1, 2, 3 are respectively mapped onto $\{b\},\{a\},\{3\}$. Where the last mapping corresponds to a deletion of 3 (which is mapped onto $m+1=3$). Consequently $m+1=3$ has two antecedents $3$ and $4=n+1$. 
  \begin{definition}[$\epsilon$-row/column stochastic matrix]
    \label{def:rowColumnStochastic}
    ~\\
    A non negative $(n+1)\times (m+1)$ matrix $X$ is called an
    $\epsilon$-row stochastic matrix iff:
    \[
      \forall i\in\{1,\dots,n\}\quad       \sum_{j=1}^{m+1}X_{i,j}=1 
    \]
    $X$ is called an $\epsilon$-column stochastic matrix iff:
      \[
        \forall j\in\{1,\dots,m\}\quad          \sum_{i=1}^{n+1}X_{i,j}=1 
      \]
  \end{definition}
  \begin{definition}[$\epsilon$-bi-stochastic matrix]
    \label{def:bistochastic}
    ~\\
    A non negative $(n+1)\times (m+1)$ matrix $X$ is called an
    $\epsilon$-bi-stochastic matrix iff:
    \[
      \left\{
        \begin{array}{ll}
          \sum_{j=1}^{m+1}X_{i,j}=1& \forall i\in\{1,\dots,n\}\\
          ~\\
          \sum_{i=1}^{n+1}X_{i,j}=1& \forall j\in\{1,\dots,m\}\\
          ~\\
          x_{n+1,m+1}=1\\
        \end{array}
        \right.
    \]
    If $X \in \{0,1\}^{(n+1)\times (m+1)}$, $X$ is called an $\epsilon$-assignment matrix and
    there is a one-to-one mapping between $\epsilon$-assignments and
    $\epsilon$-assignment matrices.
  \end{definition}

  Let us note that any $\epsilon$-bi-stochastic matrix is a bi-stochastic
  matrix on which the bi-stochastic constraints are relaxed on the last
  line and last column. So any squared bi-stochastic matrix is also an
  $\epsilon$-bi-stochastic matrix (the reverse being obviously false).

\begin{definition}{$\epsilon$-diagonal}
    \label{def:diagonal}
    ~\\
  If $A$ is a $(n+1)\times(m+1)$ matrix and $\varphi$ an
  $\epsilon$-assignment then the set
  $A_{1,\varphi(1)},\dots,A_{n+1,\varphi({n+1})}$ is called an $\epsilon$-\textbf{diagonal}
  of $A$ corresponding to $\varphi$. if $A$ is squared and $\varphi$
  is the identity, the diagonal is called the main diagonal.  
    
\end{definition}
Note that
  $a_{1,\varphi(1)},\dots,a_{n,\varphi({n})}$ is a sequence (as
  $\varphi(i)$ is unique for $i\in\{1,\dots,n\}$ while
  $a_{n+1,\varphi(n+1)}$ is a set). The above definition is a straightforward of the usual notion of diagonal where $\varphi$ is required to be a permutation.  In the following we will only consider $\epsilon$-diagonals of matrices which will be simply called diagonal. 
\begin{definition}{total support}
    \label{def:total_support}
    ~\\
  If $A$ is a nonnegative matrix, $A$ is said to have total support if   $A\neq 0$ and if every positive element of $A$ lies on a positive  $\epsilon$-diagonal. A nonnegative matrix that contains a positive diagonal is   said to have a support.
\end{definition}

  If $\mu$ and $\nu$ define set of indices respectively contained in
  $\{1, \dots,n+1\}$ and $\{1,\dots,m+1\}$ then :
  \begin{itemize}
  \item $A[\mu,\nu]$ denotes the sub matrix of $A$ restricted to indices $\mu$ and $\nu$,
    
  \item $A(\mu,\nu]$ denotes the sub matrix of $A$ restricted to indices not contained in $\mu$, i.e. $\{1,\dots,n+1\}-\mu,$ and to the indices contained in $\nu$,
  \item $A[\mu,\nu)$ denotes the sub matrix of $A$ restricted to indices contained in $\mu$ and not contained in $\nu$, i.e. $\{1,\dots,m+1\}-\nu$.
  
  \item $A(\mu,\nu)$ denotes the sub matrix of $A$ restricted to the
    indices not contained in $\mu$ and $\nu$.
  \end{itemize}
  %\begin{lemma}
   % if $A$ is a $(n+1)\times(m+1)$ matrix:
    %\begin{itemize}
    %\item If $A$ is an $\epsilon$-row stochastic matrix
    %and $\beta_1,\dots,\beta_m$ are the column sums of $A$, then
    %\[
    %  \Pi_{k=1}^m\beta_k\leq \left(\frac{n}{m}\right)^m
    %\]
    %with equality on if each $\beta_k=1$.
  %\end{itemize}

%\end{lemma}
%\begin{proof}
 % Using the arithmetic-geometric mean inequality:
  %\[
   % \Pi_{j=1}^m\beta_k\leq \left(\frac{1}{m}\sum_{j=1}^m\beta_j\right)^m=
    %  \left(\frac{1}{m}\sum_{j=1}^m\sum_{i=1}^{n+1}a_{ij}\right)^m
%\]
%    \end{proof}
%The following definiton is a straightforward extension of the notion of indecomposable matrix to non squared matrices. See~\cite{RI-FENNER1977} for further details. 

%\begin{definition}{Indecomposable matrix}
%\label{def:indecomposable}
%~\\
%If $A$ is a $n\times m$, nonnegative matrix, $A$ is said %to be fully indecomposable iff there exists no %permutation matrices $P$ and $Q$ $n\times n$ and $m\times %m$ such that:
%\[
%PAQ=\left(\begin{array}{cc}
%    A_{11} &A_{12}\\
%    0&A_{22}\\
%    \end{array}\right)
%\]
%where $A_{11}$ is a squared non vacuous sub matrix and %$A_{22}$ is also non vacuous. 
%\end{definition}
%\begin{definition}{Bipartite Graph of a rectangular matrix}
%\label{def:bipartite}
%~\\
%Let $A$ be a $n\times m$ non negative matrix. The bipartite graph $\mathcal{G}(A)=(V,E,\nu)$ is defined by $V=\{1,\dots,n\}\cup\{1,\dots,m\}$ and an edge $(i,j)$ belongs to $E$ iff $a_{j,i}>0$. 
%\end{definition}
\begin{definition}{Secable rectangular matrix}
\label{def:secable}
~\\

A rectangular $n\times m$ non negative matrix $A$ is said to be secable if one can find :
\begin{itemize}
    \item a partition of $\{1,\dots,n\}$ into two sets $X$ and $Y$ and
    \item a partition of $\{1,\dots,m\}$ into two sets $Z$ and $T$
\end{itemize} 
such that:
\[
A[X,T]=0 \mbox{ and } A[Y,Z] =0
\]
\begin{center}

\begin{tabular}{ccc}
&Z&T\\
\hline
X&A[X,Z]&0\\
Y&0&A[Y,T]\\
\end{tabular}
\end{center}
\end{definition}
Let us note that this notion of secable matrix is quite close from the one of block diagonal matrix. However, $A[X,T]$ is not required to be squared. 

\section{Existence and uniqueness}

\label{sec:ExistUniq}

  \begin{theorem}
    Let $A$ be a nonnegative $(n+1)\times (m+1)$ matrix such that $A[\{1,\dots,n\},\{1,\dots,m\}]$ does not contain any line or column filled with 0. A necessary
    and sufficient condition that there exists an $\epsilon$
    bi-stochastic matrix $B$ of the form $D_1AD_2$ where $D_1$ and
    $D_2$ are diagonal matrices with positive main diagonals and a
    last entry equal to $1$ is that $A$ has total support. If $B$
    exists then it is unique. Also $D_1$ and $D_2$ are unique if and
    only if $A$ is non secable.
  \end{theorem}
  \begin{proof}
    Let us suppose that $B=D_1AD_2$ and $B'=D'_1AD'_2$ are
    $\epsilon$-bi-stochastic matrices where
    $D_1=diag(x_1,\dots,x_n,1), D_2=diag(y_1,\dots,y_m,1)$ ,
    $D'_1=diag(x'_1,\dots,x'_n,1)$ and
    $D'_2=diag(y'_1,\dots,y'_m,1)$. If $p_i=\frac{x'_i}{x_i}$ and
    $q_i=\frac{y'_i}{y_i}$:

      \begin{eqnarray}
        \sum_{i=1}^{n+1}x_ia_{ij}y_j&=&1,\forall j=1,\dots,m;\label{eq:Bcol}\\
        \sum_{j=1}^{m+1}x_ia_{ij}y_j&=&1,\forall i=1,\dots,n\label{eq:Bline}\\
        \sum_{i=1}^{n+1}p_ix_ia_{ij}q_jy_j&=&1,\forall j=1,\dots,m;\label{eq:Bpcol}\\
        \sum_{j=1}^{m+1}p_ix_ia_{ij}q_jy_j&=&1,\forall i=1,\dots,n\label{eq:Bpline}
      \end{eqnarray}

    Let $E_j=\{i| a_{ij}>0\}, F_i=\{j| a_{ij}>0\}$ and put
  \[
   \alpha=\{i\in\{1,\dots,n\}|p_i=min_i~p_i=\underline{p}\},\quad \beta=\{j\in\{1,\dots,m\}|q_j=max_j~q_j=\overline{q}\}
  \]
  Let us note that since $x_{n+1}=y_{m+1}=x'_{n+1}=y'_{m+1}=1$ we have
  $p_{n+1}=q_{m+1}=1$, $\underline{p}\leq 1$ and $\overline{q}\geq 1$.
  Moreover, we have by hypothesis $E_j\cap\{1,\dots,n\}\neq \emptyset$ and $F_i\cap\{1,\dots,m\}\neq \emptyset$ for all $(i,j)\in\{1,\dots,n\}\times\{1,\dots,m\}$.
  
  Let us fist show that $\alpha=\emptyset\iff \beta=\emptyset$.

  If $\alpha=\emptyset$, then For all $i\in\{1,\dots,n\}~p_i>1$. Then
  using equation~(\ref{eq:Bpcol}), we have for any $j\in\{1,\dots,m\}$:
  \[
    1=\sum_{i=1}^{n+1}p_ix_ia_{ij}q_jy_j>q_j\sum_{i=1}^{n+1}x_ia_{ij}y_j=q_j
  \]
  Hence $q_j<1$ for all $j\in\{1,\dots,m\}$ and thus
  $\beta=\emptyset$.

  Conversely, if $\beta=\emptyset$, we have $q_j<1$ for all
  $j\in\{1,\dots,m\}$ . Using equation~(\ref{eq:Bpline}) for
  $i\in\{1,\dots,n\}$:
    \[
      1=\sum_{j=1}^{m+1}p_ix_ia_{ij}q_jy_j<p_i\sum_{j=1}^{m+1}x_ia_{ij}y_j=p_i
  \]
  Hence, $p_i>1$ for all $i\in\{1,\dots,n\}$ and $\alpha=\emptyset$.

  In this case  we consider the alternative definitions for $\alpha$ and $\beta$:
  \[
    \left\{
    \begin{array}{lll}
      \alpha&=&\{i\in\{1,\dots,n\}|p_i=max_i~p_i=\overline{p}\},\\
      \beta&=&\{j\in\{1,\dots,m\}|q_j=min_j~q_j=\underline{q}\}\\
    \end{array}\right.
  \]
  Since $p_i>1$ for all $i\in \{1,\dots,n\}$ and $q_j<1$ for all $j\in \{1,\dots,m\}, \alpha$ and $\beta$ are non empty.

  Using initial definitions for $\alpha$ and $\beta$, let us thus consider $i_0\in \alpha$ and $j_0\in \beta$. Then
  using~(\ref{eq:Bpcol}):
  \[
    q_{j_0}=\frac{1}{\sum_{i=1}^{n+1}p_ix_ia_{ij_0}y_{j_0}}\leq \frac{1}{p_{i_0}\sum_{i=1}^{n+1}x_ia_{ij_0}y_{j_0}}=p_{i_0}^{-1}
  \]
  where the last equality comes from~(\ref{eq:Bcol}). Similarly, using~(\ref{eq:Bpline}):
  \[
    p_{i_0}=\frac{1}{\sum_{j=1}^{m+1}x_ia_{ij}q_jy_j}\geq
    \frac{1}{q_{j_0}\sum_{j=1}^{m+1}x_ia_{ij}y_j}=q_{j_0}^{-1}
  \]
  where the last equality comes from~(\ref{eq:Bline}). Whence
  $q_{j_0}=p_{i_0}^{-1}=\underline{p}^{-1}$. But in this case, we have using (\ref{eq:Bpcol}):
  \[
    \sum_{i=1}^{n+1}p_ix_ia_{ij_0}q_{j_0}y_{j_0}=
    \sum_{i\in E_{j_0}}p_ix_ia_{ij_0}q_{j_0}y_{j_0}=
    \sum_{i\in E_{j_0}}\frac{p_i}{\underline{p}}x_ia_{ij_0}y_{j_0}=1
  \]
  This last equality is compatible with~(\ref{eq:Bcol}) only if
  $p_i=\underline{p}$ for all $i\in E_{j_0}$.
  Dropping sub indices, we have for all $j \in \beta$ and all $i\in E_j$ $p_i=\underline{p}$. Thus
  \[
    \bigcup_{j\in \beta} E_j\subseteq \alpha\cup\{n+1\}.
  \]
  Hence if $j\in \beta$ and $i\not\in \alpha\cup\{n+1\}, i\not\in \bigcup_{j\in \beta}
  E_j$. So $a_{ij}=0$. More concisely, we have: $A(\alpha\cup\{n+1\},\beta]=0$.

  In the same way, $p_{i_0}=q_{j_0}^{-1}=\overline{q}^{-1}$ implies using (\ref{eq:Bpline}):
  \[
    \sum_{j=1}^{m+1}p_{i_0}x_{i_0}a_{i_0j}q_jy_j=  \sum_{j\in F_{i_0}}\frac{q_j}{\overline{q}}  x_{i_0}a_{i_0j}y_j=1
  \]
  which is compatible with (\ref{eq:Bline}) only if $q_j=\overline{q}$
  for all $j\in F_{i_0}$. Thus for all $i\in \alpha$ and for all $j\in F_i$
  we have $q_j=\overline{q}$. Thus
  \[
    \bigcup_{i\in \alpha}F_i\subset \beta\cup\{m+1\} \mbox{ and } A[\alpha,\beta\cup\{m+1\})=0
  \]
  On $\alpha\times \beta$ we have $p_iq_j=\underline{p}\overline{q}=1$. Thus:
  \[
    a_{ij}=\frac{b_{ij}}{x_iy_j}=\frac{b'_{ij}}{x'_iy'_j}\Rightarrow p_iq_jb_{ij}=b'_{ij}\Rightarrow b_{ij}=b'_{ij}
  \]
  Hence $B[\alpha,\beta]=B'[\alpha,\beta]$

  Moreover, for any $j\in \beta$ we have:
  \[
    \sum_{i=1}^{n+1}b_{i,j}=\sum_{i=1}^{n+1}x_ia_{i,j}y_{j}=\sum_{i\in E_j\subset\alpha\cup\{n+1\}}x_ia_{i,j}y_{j}=\sum_{i\in \alpha\cup\{n+1\}}x_ia_{i,j}y_{j}=\sum_{i\in \alpha\cup\{n+1\}}b_{i,j}=1
  \]
  In the same way, we have for any
  $j\in \beta, \sum_{i\in\alpha\cup\{n+1\}}b'_{i,j}=1$. But in this case 
  using $B[\alpha,\beta]=B'[\alpha,\beta]$, we have for $j\in \beta$:
  \[
    1=\sum_{i\in\alpha\cup\{n+1\}}b'_{i,j}=\sum_{i\in \alpha}b'_{i,j}+b'_{n+1,j}=\sum_{i\in \alpha}b_{ij}+b'_{n+1,j}
\]
Thus $1-b_{n+1,j}+b'_{n+1,j}=1$ which induces $b_{n+1,j}=b'_{n+1,j}$
which imposes $\overline{q}=1$. Indeed, since $j\in \beta$, we have $q_j=\overline{q}$ and
$q_jb_{n+1,j}=\overline{q}b_{n+1,j}=b'_{n+1,j}$.

In the same way for
$i\in \alpha$:
  \[
    \sum_{j=1}^{m+1}b_{ij}=\sum_{j=1}^{m+1}x_ia_{i,j}y_{j}=\sum_{j\in F_i\subset\beta\cup\{m+1\}}x_ia_{i,j}y_{j}=\sum_{j\in \beta\cup\{m+1\}}x_ia_{i,j}y_{j}=\sum_{j\in \beta\cup\{m+1\}}b_{i,j}=1
  \]
  and we have in the same way for $i\in \alpha$, $\sum_{j\in \beta\cup\{m+1\}} b'_{i,j}=1$. We then obtain for $i\in\alpha$:
  \[
    \sum_{j\in \beta\cup\{m+1\}} b'_{i,j}=\sum_{j\in \beta} b'_{i,j}+b'_{i,m+1}=\sum_{j\in \beta} b_{i,j}+b'_{i,m+1}
\]
  Using the previous equality $1-b_{i,m+1}+b'_{i,m+1}=1$, hence
  $b_{i,m+1}=b'_{i,m+1}$ and $\underline{p}=1$ (since
  $\underline{p}b_{i,n+1}= b'_{i,n+1})$.

  Thus
  $B[\alpha\cup\{n+1\},\beta\cup\{m+1\}]=B'[\alpha\cup\{n+1\},\beta\cup\{m+1\}]$
  is an $\epsilon$ bi-stochastic matrix (where $n+1$ and $m+1$ plays
  the role of the last row and column respectively).
    
    Let us briefly show that $\{1,\dots,n\}-\alpha$ and $\{1,\dots,m\}-\beta$ are simultaneously empty or non empty. Let us fist suppose that $\alpha=\{1,\dots,n\}$ and let us consider $j\not\in\beta\cup\{m+1\}$. Since $E_j\cap\{1,\dots,n\}\neq \emptyset$ by hypothesis, it exists $i\in\{1,\dots,n\}=\alpha$ such that $j\in F_i$. But since $i\in \alpha$, we have $F_i\subset \beta\cup\{m+1\}$ and thus  a contradiction.  In the same way, if $\beta=\{1,\dots,m\}$, let us consider $i\not\in\alpha\cup\{n+1\}$. Since $F_i\cap\{1,\dots,m\}\neq \emptyset$, it exists $j\in \{1,\dots,m\}=\beta$ such that $i\in E_j\subset\alpha\cup \{n+1\}$. Again a contradiction. 
    
  If $A$ is non secable the configuration where both $\{1,\dots,n\}-\alpha$ and $\{1,\dots,m\}-\beta$ are non empty correspond to a partition of $\{1,\dots,n+1\}$ into $\alpha\cup\{n+1\}$ and its complementary and a partition of $\{1,\dots,m+1\}$ into $\beta\cup \{m+1\}$ and its complementary with no connections between $\alpha\cup\{n+1\}$ and the complementary of $\beta\cup\{m+1\}$ nor any connection between $\beta\cup\{m+1\}$ and the complementary of $\alpha\cup\{n+1\}$ (Figure~\ref{fig:Bdecomp}). Such a decomposition being refused, we have $\alpha=\{1,\dots,n\}$ and $\beta=\{1,\dots,m\}$ 
  Hence
  $A[\alpha\cup\{n+1\},\beta\cup\{m+1\}]=A$ and $D_1AD_2=D'_1AD'_2$ and
  $D_1$ and $D_2$ are unique ($\underline{p}=\overline{q}=p_{n+1}=q_{m+1}=1$).

  If the non secable property of $A$ does not hold and $A(\alpha\cup\{n+1\},\beta]$ and $A[\alpha,\beta\cup\{m+1\})$
  exist, $B(\alpha,\beta)$ and $B'(\alpha,\beta)$ exist, include the
  row $n+1$ and the column $m+1$, are $\epsilon$ bi-stochastic matrices
  and have a size lower than the one of $A$. Furthermore,
  $B(\alpha,\beta)=D"_1A(\alpha,\beta)D''_2$ and
  $B'(\alpha,\beta)=D'''_1A(\alpha,\beta)D'''_2$ where $D''_i$ and
  $D'''_j$ have like $D_i$ and $D'_j$ (from which they are derived) a
  positive main diagonal with a $1$ at last position. The argument may
  be repeated on these submatrices until $D_1AD_2=D'_1AD'_2$ is
  established.  Given that $B(\alpha,\beta)=B'(\alpha,\beta)$, we
  already know that
  $B[\alpha\cup\{n+1\},\beta\cup\{m+1\}]=B'[\alpha\cup\{n+1\},\beta\cup\{m+1\}]$
  and that $A$ and hence $B$ and $B'$ are zeros elsewhere
  (Figure~\ref{fig:Bdecomp}). Hence $B$  is equal to $B'$.  Note however, that since
  $\alpha\cup\{n+1\}\neq \{1,\dots,n+1\}$ (and the same for $\beta$)
  $D_1$ and $D_2$ are no longer unique.
  \begin{figure}
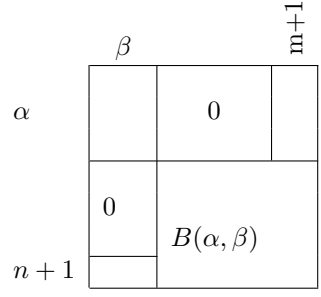

    \centering
    \[
    \begin{array}{l|cc|ccc|c|}
      \multicolumn{1}{c}{}&\multicolumn{2}{c}{\beta}&\multicolumn{3}{c}{}&\multicolumn{1}{l}{\rotatebox{90}{m+1}}\\
      \cline{2-7}
      &&&&&&\\
      \alpha& & &\multicolumn{3}{c|}{0} &\\
                          &&&&&&\\
            \cline{2-7}
                          &  & &\multicolumn{3}{c}{} &\\
                          &0& &\multicolumn{3}{c}{} &\\
                          &  & &\multicolumn{3}{c}{B(\alpha,\beta)} &\\
      \cline{2-3}
                       n+1   & & &\multicolumn{3}{c}{} &\\
            \cline{2-7}
    \end{array}
  \]  
  \caption{Decomposition of matrix $B$.}
  \label{fig:Bdecomp}
\end{figure}

\end{proof}
\section{A constructive algorithm}
\label{sec:construct}

For any $i\in \{1,\dots,n\}$ and any $j\in \{1,\dots,m\}$ let us consider the series $(x_{i,p})_{p\in\mathbb{N}}$ and $(y_{j,p})_{p\in\mathbb{N}}$ defined as follows:
\[
  \left\{
    \begin{array}{llcllcl}
      \forall i\in\{1,\dots,n\}&x_{i,0}&=&\left(\sum_{j=1}^{m+1}a_{ij}\right)^{-1}&x_{i,p+1}&=&\chi_{i,p}^{-1}x_{i,p}\\
      \forall j\in\{1,\dots,m\}&y_{j,0}&=&1&y_{j,p+1}&=&
      \gamma_{j,p}^{-1}y_{j,p}\\
    \end{array}
    \right.
  \]
  Moreover we also define:
  \[
    \left\{
      \begin{array}{llcl}
        \forall i\in\{1,\dots,n\}&\chi_{i,p}&=&\sum_{j=1}^{m+1}x_{i,p}a_{i,j}y_{j,p}=\sum_{j=1}^{m+1}\gamma_{j,p-1}^{-1}x_{i,p}a_{i,j}y_{j,p-1}\\
        \\
        \forall j\in \{1,\dots,m\}&\gamma_{j,p}&=&\sum_{i=1}^{n+1}\chi_{i,p}^{-1}x_{i,p}a_{i,j}y_{j,p}\\                                                 
      \end{array}
      \right.
    \]
    with for all $p$:
    \[
      \begin{array}{lclcl}
        x_{n+1,p}&=&y_{m+1,p}&=&1\\
        \chi_{n+1,p}&=&\gamma_{m+1,p}&=&1\\
      \end{array}
    \]
    Let us denote by $L_p$ the $(n+1)\times(m+1)$ matrix whose entries
    in $n\times(m+1)$ are equal to $(x_{i,p}a_{i,j}y_{j,p-1})$ and
    whose last row is filled with zeros but a 1 at position
    $(n+1,m+1)$.

      In the same way let us denote by $C_p$ the $(n+1)\times(m+1)$
      matrix whose entries in $(n+1)\times m$ are equal to
      $x_{i,p}a_{i,j}y_{j,p}$ and whose last column
      is filled with zeros except a $1$ at position
      $(n+1)\times(m+1)$.

      If $\chi_p\in\mathbb{R}^{n+1}$ and
      $\gamma_p\in\mathbb{R}^{m+1}$ denote the vectors encoding
      respectively $(\chi_{i,p})_{i\in\{1,\dots,n+1\}}$ and
      $(\gamma_{j,p})_{j\in\{1,\dots,m+1\}}$ we have for $p\geq 1$:
      \begin{equation}
        \left\{
          \begin{array}{lcl}
            \chi_p&=&L_p\gamma_{p-1}^{-1}\\
            \gamma_p&=&C_p^T\chi_p^{-1}\\
          \end{array}
          \right.\label{eq:alphaBeta}          
        \end{equation}
        \begin{figure}
          \centering
          \[
            \begin{array}{lcl}
            L_p&=&\left(
              \begin{array}{ccccc}
                x_{1,p}a_{1,1}y_{1,p-1}&x_{1,p}a_{1,2}y_{2,p-1}&\dots&x_{1,p}a_{1,m}y_{m,p-1}&x_{1,p}a_{1,m+1}\\
                x_{2,p}a_{2,1}y_{1,p-1}&x_{2,p}a_{2,2}y_{2,p-1}&\dots&x_{2,p}a_{2,m}y_{m,p-1}&x_{2,p}a_{2,m+1}\\
                                       &&\vdots&&\\
                x_{n,p}a_{n,1}y_{1,p-1}&x_{n,p}a_{n,2}y_{2,p-1}&\dots&x_{n,p}a_{n,m}y_{m,p-1}&x_{n,p}a_{n,m+1}\\
                0&0&\dots&0&1\\
              \end{array}
              \right)\\
              \\
              C_p&=&\left(
                     \begin{array}{ccccc}
                       x_{1,p}a_{1,1}y_{1,p}&x_{1,p}a_{1,2}y_{2,p}&\dots&x_{1,p}a_{1,m}y_{m,p}&0\\
                       x_{2,p}a_{2,1}y_{1,p}&x_{2,p}a_{2,2}y_{2,p}&\dots&x_{2,p}a_{2,m}y_{m,p}&0\\
                                            &&\vdots&&\\
                       x_{n,p}a_{n,1}y_{1,p}&x_{n,p}a_{n,2}y_{2,p}&\dots&x_{n,p}a_{n,m}y_{m,p}&0\\
                       a_{n+1,1}y_{1,p}&a_{n+1,2}y_{2,p}&\dots&a_{n+1,m}y_{m,p}&1\\
                     \end{array}
              \right)\\
              \\
              C_p^T&=&\left(
                       \begin{array}{ccccc}
                         x_{1,p}a_{1,1}y_{1,p}&x_{2,p}a_{2,1}y_{1,p}& \dots&x_{n,p}a_{n,1}y_{1,p}&a_{n+1,1}y_{1,p}\\
                         x_{1,p}a_{1,2}y_{2,p}&x_{2,p}a_{2,2}y_{2,p}&\dots&x_{n,p}a_{n,2}y_{2,p}&a_{n+1,2}y_{2,p}\\
                                              &&\vdots&&\\
                         x_{1,p}a_{1,m}y_{m,p}&  x_{2,p}a_{2,m}y_{m,p}&\dots&x_{n,p}a_{n,m}y_{m,p}&a_{n+1,m}y_{m,p}\\
                         0&0&\dots&0&1\\
                       \end{array}
                       \right)
            \end{array}
          \]
          
          \caption{$L_p$ and $C_p$ matrices.}
          \label{fig:LpCp}
        \end{figure}
        $L_p$ is row stochastic. Indeed, for any $i\in\{1,\dots,n\}$ and $p\geq 1$:
        \[
          \begin{array}{lcl}
            \sum_{j=1}^{m+1}(L_p)_{i,j}&=&\sum_{j=1}^{m+1}x_{i,p}a_{i,j}y_{j,p-1}\\
                                       &=&\sum_{j=1}^{m+1}\chi_{i,p-1}^{-1}x_{i,p-1}a_{i,j}y_{j,p-1}\\
                                       &=&\chi_{i,p-1}^{-1}\sum_{j=1}^{m+1}x_{i,p-1}a_{i,j}y_{j,p-1}\\
            &=&1\\
          \end{array}
        \]
        and the last line of $L_p$ contains a single entry equal to $1$.

        Moreover, $C_p$ is column stochastic for $p\geq 1$. Indeed for
        each $j\in\{1,\dots,m\}$:
        \[
          \begin{array}{lcl}
            \sum_{i=1}^{n+1}(C_p)_{i,j}&=&\sum_{i=1}^{n+1}x_{i,p}a_{i,j}y_{j,p}\\
                                       &=&\sum_{i=1}^{n+1}\chi_{i,p-1}^{-1}\gamma_{j,p-1}^{-1}x_{i,p-1}a_{i,j}y_{j,p-1}\\
                                       &=&\gamma_{j,p-1}^{-1}\sum_{i=1}^{n+1}\chi_{i,p-1}^{-1}x_{i,p-1}a_{i,j}y_{j,p-1}\\
            &=&1\\
          \end{array}
        \]
        Note that $C_0$ is not column stochastic. One noticeable
        effect of this negative property is that $\gamma_0\neq 1$ while
        $\chi_0=1$.
        
        Moreover the last column contains a single positive entry
        equal to $1$. Hence $C_p^T$ is row stochastic and
        equation~\ref{eq:alphaBeta} involves two row stochastic
        matrices.

        Combining both equations of~\ref{eq:alphaBeta} we have:
        \[
          \chi_p=L_p\left(C_{p-1}^T\chi_{p-1}^{-1}\right)^{-1}
        \]
        where the inverse notation applied to a vector denotes the element-wise inverse operation.

        Since $L_p$ is row stochastic, we have:
        \[
            \chi_p-1 =L_p\left[\left(C_{p-1}^T\chi_{p-1}^{-1}\right)^{-1}-1\right]
          \]
          Using $\frac{1}{x}-1=\frac{1-x}{x}$ we obtain:
          \[
            \chi_p-1 =L_p\left[\left(1-C_{p-1}^T\chi_{p-1}^{-1}\right)\odot \left(C_{p-1}^T\chi_{p-1}^{-1}\right)^{-1}\right]
          \]
          where $\odot$ is the element-wise product also known as Hadamard product. Since, for $p\geq 2$,
          $C_{p-1}^T$ is row stochastic we have $C_{p-1}^T1=1$ and
          thus:
          \[
            \begin{array}{lcl}
              \chi_p-1 &=&L_p\left[\left(C_{p-1}^T\left(1-\chi_{p-1}^{-1}\right)\right)\odot \left(C_{p-1}^T\chi_{p-1}^{-1}\right)^{-1}\right]\\
              \\
              &=&L_pdiag\left(C_{p-1}^T\chi_{p-1}^{-1}\right)^{-1}C_{p-1}^T\left(1-\chi_{p-1}^{-1}\right)\\
              
            \end{array}
          \]
          Using $1-\frac{1}{x}=\frac{x-1}{x}$ we obtain:
          \[
            \begin{array}{lcl}
              \chi_p-1  &=&L_pdiag\left(C_{p-1}^T\chi_{p-1}^{-1}\right)^{-1}C_{p-1}^T\left(\chi_{p-1}-1\right)\odot\chi_{p-1}^{-1}\\
                          &=&L_pdiag\left(C_{p-1}^T\chi_{p-1}^{-1}\right)^{-1}C_{p-1}^Tdiag(\chi_{p-1})^{-1}\left(\chi_{p-1}-1\right)\\
                          &=&L_pdiag\left(\gamma_{p-1}\right)^{-1}C_{p-1}^Tdiag(\chi_{p-1})^{-1}\left(\chi_{p-1}-1\right)\\
            \end{array}
          \]
          Moreover the left and right multiplications of $C_{p-1}^T$
          by diagonal matrices is equivalent to a multiplications of
          its lines by $\gamma_{p-1}^{-1}$ and its columns by
          $\chi_{p-1}^{-1}$. More precisely we have:
          \[
            \begin{array}{l}
            diag\left(\gamma_{p-1}\right)^{-1}C_{p-1}^Tdiag(\chi_{p-1})^{-1}=\\
          \left(
                       \begin{array}{cccc}
                         \chi_{1,p-1}^{-1}\gamma_{1,p-1}^{-1}x_{1,p-1}a_{1,1}y_{1,p-1}& \dots&\chi_{n,p-1}^{-1}\gamma_{1,p-1}^{-1}x_{n,p-1}a_{n,1}y_{1,p-1}&\gamma_{1,p-1}^{-1}a_{n+1,1}y_{1,p}\\
                         \chi_{1,p-1}^{-1}\gamma_{2,p-1}^{-1}x_{1,p-1}a_{1,2}y_{2,p-1}&\dots&\chi_{n,p-1}^{-1}\gamma_{2,p-1}^{-1}x_{n,p-1}a_{n,2}y_{2,p-1}&\gamma_{2,p-1}^{-1}a_{n+1,2}y_{2,p}\\
                                              &\vdots&&\\
                         \chi_{1,p-1}^{-1}\gamma_{m,p-1}^{-1}x_{1,p-1}a_{1,m}y_{m,p-1}&\dots&\chi_{n,p-1}^{-1}\gamma_{m,p-1}^{-1}x_{n,p-1}a_{n,m}y_{m,p-1}&\gamma_{m,p-1}^{-1}a_{n+1,m}y_{m,p}\\
                         0&\dots&0&1\\                         
                       \end{array}
              \right)\\
              \\
              =
          \left(
                       \begin{array}{ccccc}
                         x_{1,p}a_{1,1}y_{1,p}&x_{2,p}a_{2,1}y_{1,p}& \dots&x_{n,p}a_{n,1}y_{1,p}&a_{n+1,1}y_{1,p}\\
                         x_{1,p}a_{1,2}y_{2,p}&x_{2,p}a_{2,2}y_{2,p}&\dots&x_{n,p}a_{n,2}y_{2,p}&a_{n+1,2}y_{2,p}\\
                                              &&\vdots&&\\
                         x_{1,p}a_{1,m}y_{m,p}&  x_{2,p}a_{2,m}y_{m,p}&\dots&x_{n,p}a_{n,m}y_{m,p}&a_{n+1,m}y_{m,p}\\
                         0&0&\dots&0&1\\                         
                       \end{array}
              \right)=C_p^T
              
            \end{array}
          \]
          Hence we have:
          \begin{equation}
            \forall p\geq 2\quad             \chi_p-1=L_pC_p^T(\chi_{p-1}-1)\label{eq:recAlpha}            
          \end{equation}
          As $L_p$ and $C_p^T$ are row stochastic matrices, so is
          $L_pC_p^T$.  Moreover, $L_pC_p^T$ is a square
          $(n+1)\times (n+1)$ matrix.  Let us note that due do this
          row stochastic property the last equality is equivalent to:
          \[
            \forall p\geq 2\quad \chi_p=L_pC_p^T\chi_{p-1}
          \]
                    Examining more precisely the matrix $L_pC_p^T$ we have:
          \[
            \left\{
              \begin{array}{lcll}
                (L_pC_p^T)_{i,j}&=&x_{i,p}x_{j,p}\sum_{k=1}^ma_{i,k}a_{j,k}y_{k,p}y_{k,p-1}&\forall (i,j)\in\{1,\dots,n\}^2\\
                (L_pC_p^T)_{n+1,j}&=&0&\forall j\in\{1,\dots,n\}\\
                (L_pC_p^T)_{i,n+1}&=&x_{i,p}\sum_{k=1}^ma_{i,k}a_{n+1,k}y_{k,p}y_{k,p-1}+x_{i,p}a_{i,m+1}&\forall i\in \{1,\dots,n\}\\
                (L_pC_p^T)_{n+1,n+1}&=&1\\
              \end{array}
            \right.
          \]
          Let us note that $i$ and $j$ may be interchanged in the
          first equation above. Hence:
          \[
            \forall (i,j)\in \{1,\dots,n\}^2 \quad (L_pC_p^T)_{i,j}=(L_pC_p^T)_{j,i}
          \]

          Moreover we have for all $(i,j)\in\{1,\dots,n\}\times\{1,\dots,m\}$:
          \[
            \left\{
            \begin{array}{lclcl}
              \chi_{i,0}&=&x_{i,0}\sum_{j=1}^{m+1}a_{i,j}y_{j,0}&=&\frac{\sum_{j=1}^{m+1}a_{i,j}}{\sum_{j=1}^{m+1}a_{i,j}}=1\\
            \gamma_{j,0}&=&\sum_{i=1}^{n+1}\chi_{i,0}^{-1}x_{i,0}a_{i,j}y_{j,0}&=&\sum_{i=1}^{n+1}\frac{a_{i,j}}{\sum_{j=1}^{m+1}a_{i,j}}\\
            \end{array}
          \right.
        \]
        Since the sum of a line (or a column) of $A$ cannot be equal
        to zeros, it exists two positive numbers
        $\underline{\gamma},\overline{\gamma}$ such that:
        \[
          0<\underline{\gamma}1\leq\gamma_0\leq \overline{\gamma}1
        \]
        We have thus:
        \[
          \begin{array}{lclcl}
            0<\underline{\gamma}1&\leq&\gamma_0&\leq&\overline{\gamma}1\\
            \overline{\gamma}^{-1}1&\leq&\gamma^{-1}_0&\leq&\underline{\gamma}^{-1}1\\
            \overline{\gamma}^{-1}1&\leq&L_1\gamma^{-1}_0&\leq&\underline{\gamma}^{-1}1\\
            0<\overline{\gamma}^{-1}1&\leq&\chi_1&\leq&\underline{\gamma}^{-1}1\\
          \end{array}
        \]
        where the last inequality is deduced from the fact that
        $\chi_1=L_1\gamma^{-1}_0$, and the fact that all the entries
        of $L_1$ are non negative.

        Hence using~(\ref{eq:recAlpha}) and a basic recursion we have :
        \[
          \forall p\geq 1,\quad 0< \overline{\gamma}^{-1}1\leq\chi_p\leq \underline{\gamma}^{-1}1
        \]
        Using $\gamma_p=C_p^T\chi^{-1}_{p}$ which is row stochastic we obtain:

        \[
          \forall p\geq 1,\quad 0< \underline{\gamma}1\leq\gamma_p\leq \overline{\gamma}1
        \]

        From now on, let us suppose that all entries of the last column
        and the last line of $A$ are positive.

        Let us suppose that for some $i\in\{1,\dots,n\}$ we have
        $\lim_{p\rightarrow+\infty}x_{i,p}=+\infty$. Since we have:
        \[
          \chi_{i,p}=x_{i,p}\left[\sum_{j=1}^ma_{i,j}y_{j,p}+a_{i,m+1}\right]
        \]
        we also have $\lim_{p\rightarrow +\infty}\chi_{i,p}=+\infty$
        which is impossible since $\chi$ is bounded.
        It exists thus an upper bound $M_x$ such that:
        \[
          \forall i\in\{1,\dots,n+1\}\forall p\geq 2\quad  x_{i,p}\leq M_x
        \]
        In this case we have:
        \[
          1=y_{j,p+1}\sum_{i=1}^{n+1}a_{i,j}x_{i,p+1}\leq y_{j,p+1}M_x\sum_{i=1}^{n+1}a_{i,j}\leq y_{j,p+1}M_x\|A\|_1
          \]
          Hence we have:
          \[
            y_{j,p+1}\geq \frac{1}{M_x\|A\|_1}
            \]
        
            In the same way, let us suppose that for some
            $j\in\{1,\dots,m\}$ we have
            $\lim_{p\rightarrow+\infty}y_{j,p}=+\infty$. Since we
            have:
            \[
              \gamma_{j,p}=y_{j,p}\left[\sum_{i=1}^n\chi_{i,p}^{-1}a_{i,j}x_{i,p} +a_{n+1,j}\right]
            \]
            we also have
            $\lim_{p\rightarrow +\infty}\beta_{j,p}=+\infty$ which is
            forbidden since $\gamma$ is also upper bounded.

            So, it exists an upper bound $M_y$ such that:
            \[
              \forall j\in\{1,\dots,m+1\},\forall p\geq 2 \quad y_{j,p}\leq M_y
            \]
            In this case we have:
            \[
              1=x_{i,p+1}\sum_{j=1}^{m+1}a_{i,j}y_{j,p}\leq x_{i,p+1}M_y\sum_{j=1}^{m+1}a_{i,j}\leq x_{i,p+1}M_y\|A\|_\infty
            \]
            Hence:
            \[
              x_{i,p+1}\geq \frac{1}{M_y\|A\|_\infty}
            \]
            \begin{enumerate}
                \item  The series $(x_{i,p})_p$ and $(y_{j,p})_p$ being both
            lower bounded, all non zeros entries of $L_pC_p^T$ are
            lower bounded by a positive value. Moreover since $L_pC_p^T$ is row stochastic, all  its entries are bounded by $1$. We say that the entries of $L_pC_p^T$ are \emph{uniformly positive}. 

            \item Moreover:
            \[
              \forall i\in\{1,\dots,n\}\quad (L_pC_p)_{i,i}=x^2_{i,p}\sum_{k=1}^m(a_{i,k})^2y_{k,p}y_{k,p-1}
            \]
            Since both series $(x_{i,p})_p$ and $(y_{j,p})_p$ are
            lower bounded and that a line of $A$ cannot be equal to
            zeros, we have $(L_pC_p^T)_{i,i}>0$. Let us additionally
            note that $(L_pC_p^T)_{n+1,n+1}=1>0$.
            
            \item Let us consider the Graph $\mathcal{G}(L_pC_p^T)=(V,E)$ where $V=\{1,\dots,n+1\}$ and an edge $(i,j)$   connects node $i$ to node $j$ iff $(L_pC_p^T)_{j,i}>0$. In this case since for any $i\in\{1,\dots,n\}$, $(L_pC_p^T)_{i,n+1}\geq x_{i,p}a_{i,m+1}>0$. The node $n+1$ is adjacent to all nodes in $\{1,\dots,n\}$. Conversely, since $(LC_p^T)_{n+1,j}=0$ for all $j\in \{1,\dots,n\}$, no node $j$ is incident to $n+1$. The node $n+1$ defines  a \emph{source component}. Moreover, for any pair $(i,j)\in\{1,\dots,n\}^2$,  since $(L_pC_p^T)_{i,j}=(L_pC_p^T)_{j,i}$, node $i$ is adjacent to node $j$ and vice versa. Hence $n+1$ is the only source component of $\mathcal{G}(L_pC_p^T)$ which is thus \emph{quasi-strongly connected}. 
     
            \end{enumerate}

            Using~\cite{RI-proskurnikov2020}, Lemma 3 with T=1 we can conclude that
            $\chi$ converges towards a consensus of the form
            $c1$. Since the last entry of $\chi$ is a constant 
            equal to $1$ we have $c=1$ and:
            \[
              \lim_{p\rightarrow+\infty}\chi_p=1
            \]
            Since $C_p^T$ is row stochastic we have:
            \[
              \lim_{p\rightarrow+\infty}\gamma_p=\lim_{p\rightarrow+\infty}C_p^T\chi^{-1}_p=1
            \]
            Since $\chi_{i,p}=\frac{x_{i,p}}{x_{i,p+1}}$ converges
            towards $1$, $(x_{i,p})_p$ is a Cauchy serie in a complete
            space($\mathbb{R}^+$). Hence $(x_{i,p})_p$ converges.
            The same argument holds for $(y_{j,p})_p$.

    Let us consider the two diagonal matrices $D_{1,p}=diag(x_{1,p},\dots,x_{n+1,p})$ and $D_{2,p}=diag(y_{1,p},\dots,y_{m+1,p})$ together with the matrix $S_p=D_{1,p}AD_{2,p}$.  By construction we have for any $i\in\{1,\dots,n\}$:
    \[
    (S_p1)_i=\sum_{j=1}^{m+1}x_{i,p}a_{i,j}y_{j,p}=\chi_{i,p}
    \]
    Thus: $\lim_{p\rightarrow \infty}(S_p 1)_i=\lim_{p\rightarrow\infty}\chi_{i,p}=1$. The matrix $S_p$ converges thus towards an $\epsilon$ row stochastic matrix. 
    
    Moreover, for any $j\in\{1,\dots,m\}$:
    \[
    (S_p^T1)_j=\sum_{i=1}^{n+1}x_{i,p}a_{i,j}y_{j,p}=\sum_{i=1}^{n+1}x_{i,p+1}a_{i,j}y_{j,p}+(x_{i,p}-x_{i,p+1})a_{i,j}y_{j,p}
    \]
     Since $x_{i,p}$ converges for any $i\in\{1,\dots,n+1\}$, $y_{j,p}$ is bounded, and the above sums are finite, it exists for any $\eta>0$ a value $p_0$ such that for any $p\geq p_0$, we have :
     \[
     \left|\sum_{i=1}^{n+1}x_{i,p}a_{i,j}y_{j,p}-\sum_{i=1}^{n+1}x_{i,p+1}a_{i,j}y_{j,p}\right|<\eta
     \]
     Both sums converge (or diverge) thus toward a same value. Moreover:
     \[
     \sum_{i=1}^{n+1}x_{i,p+1}a_{i,j}y_{j,p}=\gamma_{j,p}
     \]
     We have thus: $\lim_{p\rightarrow \infty}(S_p^T1)_j=\lim_{p\rightarrow\infty}\gamma_{j,p}=1$. The matrix $S_p$ converges thus toward  an $\epsilon$ column stochastic matrix and hence an $\epsilon$ bi-stochastic matrix. 
     
   \section{Two iterative algorithms}
   \label{sec:algo}
   
     \begin{figure}
         \centering
         \begin{lstlisting}[firstnumber=1]
         def sinkhorn_D1D2(S,nb_iter,eps):
            ones_n = torch.ones(S.shape[0],device=S.device)
            ones_m = torch.ones(S.shape[1],device=S.device)
            c=ones_m
            converged=False
            i=0
            while i <=nb_iter and not converged:
                xp=1.0/(S@c)
                xp[-1]=1.0
                if i>=1:
                    # computation of ||x_{p+1}/x_p-1||$
                    norm_x=tl.norm(x/xp-torch.ones_like(x/xp),ord=float('inf'))
                x=xp
                yp=1.0/(S.T@r)
                yp[-1]=1.0
                # computation of ||y_{p+1}/y_p-1||$
                norm_y=tl.norm(y/yp-torch.ones_like(y/yp),ord=float('inf'))
                y=yp
                if i>=1:
                    converged= (norm_x <= eps) and (norm_y<=eps)
                i+=1

        return torch.diag(x)@S@torch.diag(y)
         \end{lstlisting}
         \caption{The code corresponding to the construction scheme described in Section~\protect\ref{sec:construct}}
         \label{fig:sinkhornD1D2}
     \end{figure}
     
     \begin{figure}
         \centering
         \begin{lstlisting}
         def sinkhorn_Sp(S,nb_iter,eps):
            ones_n = torch.ones(S.shape[0],device=S.device)
            ones_m = torch.ones(S.shape[1],device=S.device)
    
            i=1
            Sp=S
            while i<=nb_iter and not converged:
                D=torch.diag(1.0/(Sp@ones_m))
                D[D.shape[0]-1,D.shape[1]-1] =1.0
                Sp1 = D@Sp
                norm_col=tl.norm((ones_n@Sk1-ones_m)[0:-1],ord=float('inf'))
                D=torch.diag(1.0/(ones_n@Sp1))
                D[D.shape[0]-1,D.shape[1]-1]=1.0
                Sp = Sp1@D
                norm_line=tl.norm((Sp@ones_m-ones_n)[0:-1],ord=float('inf'))
                converged=(norm_col <= eps) and (norm_line<=eps)
                i+=1
            D=torch.diag(1.0/(Sp@ones_m))
            D[D.shape[0]-1,D.shape[1]-1] =1.0
            Sp1[0:n,:]=(D@Sp)[0:n,:]
            Sp1[-1,0:m]=ones_m[0:m]-torch.sum((D@Sp)[0:n,0:m],dim=0)


            return Sp1
         \end{lstlisting}
         \caption{A version of our computation of an $\epsilon$ assignment matrix which updates directly the matrix $S_p$}
         \label{fig:sinkhornSp}
     \end{figure}
     %           Sp = S
%            for i in range(nb_iter):
%                D=torch.diag(1.0/(Sp@ones_m))
%                D[D.shape[0]-1,D.shape[1]-1]=1.0
%                Sp1 = D@Sp
%                D=torch.diag(1.0/(ones_n@Sp1))
%                D[D.shape[0]-1,D.shape[1]-1]=1.0
%                Sp = Sp1@D
        
%            return Sk
     The code (in python) corresponding to the construction of matrices $D_{1,p}$ and $D_{2,p}$ is provided in Figure~\ref{fig:sinkhornD1D2}. You may note the fact that we set $x_{n+1}$ and $y_{m+1}$ to $1$ respectively on line 9 and 15. This point together with the use of rectangular matrices is the main difference between this algorithm and the "classical" Sinkhorn algorithm. The convergence criterion which allows to avoid to loop up to the maximum number of iterations is based on the fact that both $\chi_p$ and $\gamma_p$ converge toward a vector of $1$.

     An equivalent code computing directly the matrix $S_p$ is provided in Figure~\ref{fig:sinkhornSp}. In this case the setting of $x_{n+1}$ and $y_{m+1}$ to $1$ is performed on line 9 and 19. The stopping criterion is based on the computation of the distance of the current matrix to the set of $\epsilon$ assignment matrices. To do so, we compute the distance between  the vector of $1$ and $S_p^T1$  after each row normalization(line 11). In the same way, we compute the distance between a vector of $1$ and $S_p1$ after each column normalization (line 15).  After convergence, we apply a last row normalization before setting the last line of our $\epsilon$ assignment matrix to the complement to $1$ of each column. 
     
\section{From similarity to cost matrices and vice versa}
\label{sec:sim_to_cost}

The Sinkhorn algorithm is well known for providing an approximation of the Linear Sum Assignment Problem (Section~\ref{sec:xp}) which can be formulated as:
\[
\max_{X}\sum_{i=1}^n\sum_{j=1}^ns_{i,j}x_{i,j}
\]
where $S=(s_{i,j})$ is our similarity matrix and $X=(x_{i,j})$ is taken over all bi stochastic matrices. The optimal solution being a permutation matrix, hence a binary matrix. 

This maximization problem may be translated into a minimization problem by considering the matrix $c \mathds{1}_{n\times n}-S$, where $\mathds{1}_{n\times n}$ is a $n\times n$ matrix filled of $1$ and $c$ is a positive constant greater than all values of $S$.  We have indeed:
\[
\begin{array}{lcl}
  \sum_{i=1}^n\sum_{j=1}^n(c-s_{i,j})x_{i,j}&=& \sum_{i=1}^n\sum_{j=1}^ncx_{i,j} -\sum_{i=1}^n\sum_{j=1}^ns_{i,j}x_{i,j}\\
     &=& c\sum_{i=1}^n\sum_{j=1}^nx_{i,j} -\sum_{i=1}^n\sum_{j=1}^ns_{i,j}x_{i,j}\\
     &=&c\sum_{i=1}^n1 -\sum_{i=1}^n\sum_{j=1}^ns_{i,j}x_{i,j}\\
     &=& cn-\sum_{i=1}^n\sum_{j=1}^ns_{i,j}x_{i,j}\\
\end{array}
\]
Hence $c$ and $n$ being constant, minimize $\sum_{i=1}^n\sum_{j=1}^n(c-s_{i,j})x_{i,j}$ is equivalent to maximize $\sum_{i=1}^n\sum_{j=1}^ns_{i,j}x_{i,j}$. The matrix $c\mathds{1}_{n\times n}-S$ is usually interpreted as a cost matrix. This last point is important if one wants to compare the Sinkhorn algorithm to an optimal Hungarian algorithm which performs a minimization of costs instead of a maximization of similarities. 

As stated in Section~\ref{sec:intro}, our algorithms being an extension of the Sinkhorn algorithm we expect them to converge to :
\[
\max_{x}\sum_{i=1}^{n+1}\sum_{j=1}^{m+1}s_{i,j}x_{i,j}
\]
where $X=(x_{i,j})$ is taken over all $\epsilon$-bi stochastic matrices.  However, the transformation of this maximization of similarities into a minimization of costs, is slightly more complex in the case of  $\epsilon$ assignment matrices.  To do so, let us consider a $(n+1)\times (m+1)$ matrix $C=(c_{i,j})$ with:
\[
c_{i,j}=\left\{
\begin{array}{lll}
2c&\mbox{ if}& i\leq n\wedge j\leq n\\
c_{lr}&\mbox{ if}& i=n+1\wedge j\leq m\\
c_{lc}&\mbox{ if} &i\leq n\wedge j=m+1\\
0&\mbox{ if} &i=n+1\wedge j=m+1\\
\end{array}\right.
\]
where $c,c_{lr},c_{lc}$ are three positive constants.  Considering the cost matrix $C-S$ we have:
\[
  \sum_{i=1}^{n+1}\sum_{j=1}^{m+1}(c_{i,j}-s_{i,j})x_{i,j}= \sum_{i=1}^{n+1}\sum_{j=1}^{m+1}c_{i,j}x_{i,j} -\sum_{i=1}^{n+1}\sum_{j=1}^{m+1}s_{i,j}x_{i,j}
\]
\begin{description}
    \item[If $c_{lc}=2c$ and $c_{lr}=0$:] We have:
  \[
  \begin{array}{lcl}
        \sum_{i=1}^{n+1}\sum_{j=1}^{m+1}c_{i,j}x_{i,j}&=& \sum_{i=1}^{n}\sum_{j=1}^{m+1}2cx_{i,j}\\
        &=&2c\sum_{i=1}^{n}1\\
        &=&2cn\\
  \end{array}
\]  
\item[If $c_{lr}=2c$ and $c_{lc}=0$:] We have:
  \[
  \begin{array}{lcl}
        \sum_{i=1}^{n+1}\sum_{j=1}^{m+1}c_{i,j}x_{i,j}&=& \sum_{j=1}^{m}\sum_{i=1}^{n+1}2cx_{i,j}\\
        &=&2c\sum_{j=1}^{m}1\\
        &=&2cm\\
  \end{array}
\]  
\item[If $c_{lr}=c$ and $c_{lc}=c$:] We have:

\[
  \begin{array}{lcl}
          \sum_{i=1}^{n+1}\sum_{j=1}^{m+1}c_{i,j}x_{i,j}&=&2c\sum_{i=1}^n\sum_{j=1}^{m}x_{i,j}+c\sum_{i=1}^nx_{i,m+1}+c\sum_{j=1}^{m}x_{n+1,j}\\
        &=&c\sum_{i=1}^n\sum_{j=1}^{m+1}x_{i,j}+c\sum_{j=1}^m\sum_{i=1}^{n+1}x_{i,j}\\
        &=& cn+cm\\
  \end{array}
\]
\end{description}
In all cases we have thus:
\begin{equation}
\sum_{i=1}^{n+1}\sum_{j=1}^{m+1}(c_{i,j}-s_{i,j})x_{i,j}=Q-\sum_{i=1}^{n+1}\sum_{j=1}^{m+1}s_{i,j}x_{i,j}    
\label{eq:min_max}
\end{equation}
where $Q=cn$ if $c_{lc}=2c$ and $c_{lr}=0$, $Q=cm$ if $c_{lr}=2c$ and $c_{lc}=0$ and finally $Q=cn+cm$ if $c_{lr}=c_{lc}=c$.
The minimization of the left part of equation~\ref{eq:min_max} (minimization of costs) is thus equivalent to a maximization of the similarities. 

Let us note that the trivial solution consisting to take $C=c\mathds{1}_{(n+1)\times(m+1)}$ does not provide an equivalence between both problems since additional terms related either to the last column or the last row forbid to state that one problem is equal to a constant minus the other problem.  Moreover, the last solution ( $c_{lr}=c$ and $c_{lc}=c$) is the only one allowing to ensure that all coefficients of the similarity matrix are positive when  transforming a cost matrix into a similarity matrix.

\section{Experiments}
\label{sec:xp}
We  proposed  in Section~\ref{sec:construct} two algorithms converging toward an unique solution if the conditions defined in Section~\ref{sec:ExistUniq} are satisfied. The aim of this section is to measure experimentally the convergence of our algorithms toward a solution maximizing :
\[
\sum_{i=1}^{n+1}\sum_{j=1}^{m+1}s_{i,j}x_{i,j}
\]
over all $\epsilon$ bi stochastic matrices $X$. Where $S$ is the input matrix. Such a problem is called a Linear Sum Assignment Problem with Edition (LSAPE). 

\subsection{Deviation of the Sinkhorn algorithm from the optimal solution}

\begin{figure}
    \centering
    \includegraphics{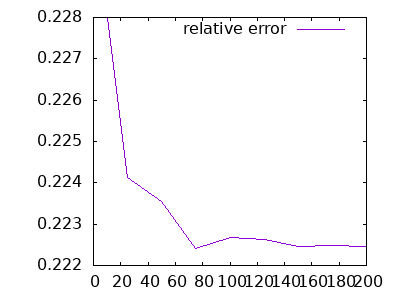}
    \caption{Relative Error between the results provided by the Sinkhorn Algorithm and the optimal solution provided by an Hungarian method.}
    \label{fig:sinkhorn_rel_error}
\end{figure}

Sinkhorn algorithm provides an approximate solution to the well known Linear Sum Assignment Problem (LSAP):
\[
\max_{X}\sum_{i=1}^{n}\sum_{j=1}^{m}s_{i,j}x_{i,j}
\]
where $X$ is taken over the set of bi stochastic matrices. From a certain point of view, LSAP  may be considered as a restriction of  the LSAPE  with squared matrices and no deletions/insertions.  Let us first evaluate the error induced by the use of the Sinkhorn algorithm. To do this, we define  matrices filled by random number in the interval  $[1,2]$. For each  matrix size we compute $100$ matrices and compute for each matrix both the solution produced by the Sinkhorn algorithm and the optimal one produced by an Hungarian algorithm. The results of this experiment are displayed in Figure~\ref{fig:sinkhorn_rel_error} for $n\in \{10,\dots,200\}$.

Considering Figure~\ref{fig:sinkhorn_rel_error}, the error is approximately constant for all sizes of matrices lies between  $20\%$ and $23\%$.

\subsection{Deviation of our algorithms from the optimal solution}

In order to test our algorithm we use the same kind of random matrices but with a specific procedure for the last row and column which encode respectively the affinity of each element toward  insertions and deletions: 

\begin{equation}
    S_{i,j}\coloneqq\left\lbrace\begin{smallmatrix*}[l]\rand()+1&\text{if}~i<n\,\wedge\,j<m\\0&\text{if}~i=n\,\wedge\,j=m\\h\rand()&\text{\,else}\end{smallmatrix*}\right.
\end{equation}

For $h\leq 0.5$ we can insure that for any $(i,j)\in\{1,\dots,n\}\times\{1,\dots,m\}$, $s_{i,j}\geq s_{n+1,j}+s_{j,m+1}$. In other terms we always get a greater sum by substituting $i$ onto  $j$ than by deleting $i$ and then inserting $j$. Conversely, if $s_{i,j}< s_{n+1,j}+s_{j,m+1}$ the substitution of $i$ onto $j$ will never be part of an optimal $\epsilon$-assignment since this operation can be replaced, with a greater value of the sum, by the removal of $i$ and the insertion of $j$.

\begin{figure}
    \centering
    \begin{tabularx}{1.0\textwidth}{m{1cm}m{3cm}m{3cm}m{3cm}}
    & \begin{minipage}{.3\textwidth}
    \centering \%error\\
    sinkhorn\_Sp
    \end{minipage}
    &
      \begin{minipage}{.3\textwidth}
      \centering \%error\\ sinkhorn\_D1D2
      \end{minipage}
      &
      \begin{minipage}{.3\textwidth}
      \[
     \frac{error S_p-error D_1D_2}{error S_p}
     \]
     \end{minipage}\\
        $n\times n$
        &
         \includegraphics[width=.3\textwidth]{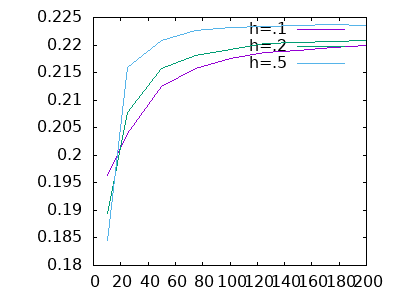}&
         \includegraphics[width=.3\textwidth]{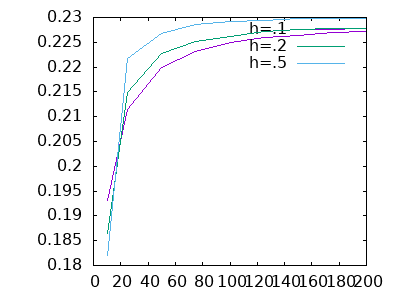}&
         \includegraphics[width=.3\textwidth]{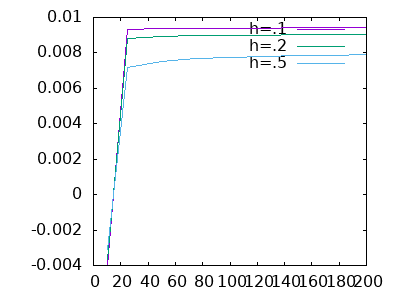}  \\
    $n\times 2n$&
    \includegraphics[width=.3\textwidth]{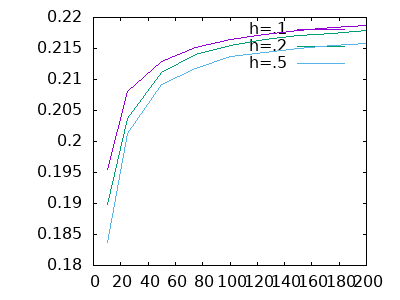}&
    \includegraphics[width=.3\textwidth]{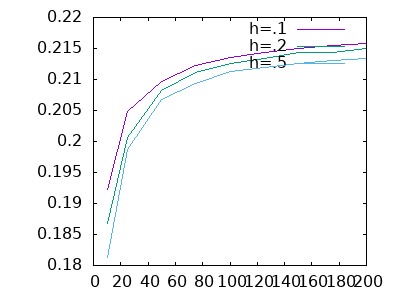}&
    \includegraphics[width=.3\textwidth]{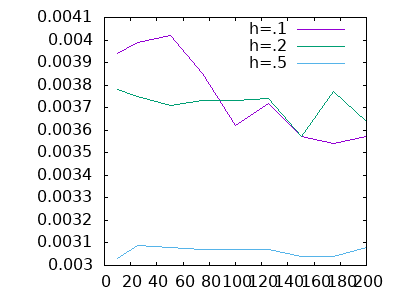}\\
    \end{tabularx}
    
    \caption{Relative errors of sinkhorn\_Sp and sinkhorn\_D1D2 according to the optimal solution. }
    \label{fig:rel_errorh.1-.5}
\end{figure}

Let us first focus on values of $h$ lower than $.5$. Figure~\ref{fig:rel_errorh.1-.5} shows on the first two columns the relative error of both sinkhorn\_Sp and sinkhorn\_D1D2 according to an optimal LSAPE algorithm~\cite{CI-Bougleux2017} for increasing sizes of the matrix. Let us note that~\cite{CI-Bougleux2017} minimizes a sum of costs. We compare both algorithms using the results of Section~\ref{sec:sim_to_cost}. For each matrix size $100$ random matrices are generated and the results are averaged for all three algorithm (sinkhorn\_Sp, sinkhorn\_D1D2 and the optimal one). Our algorithms provide an approximation of the LSAPE which is slightly above $20\%$, hence comparable with the one provided by the Sinkhorn algorithm for the LSAP problem. Interestingly, our algorithm provides better approximations for small matrix sizes while Figure~\ref{fig:sinkhorn_rel_error} suggest an opposite behavior for the Sinkhorn algorithm. 

The last column of Figure~\ref{fig:rel_errorh.1-.5} allows to compare more precisely sinkhorn\_Sp and sinkhorn\_D1D2. Both algorithms seems to be equivalent since the relative error between both methods do not exceed $1\%$. 

\begin{figure}
    \centering
    \begin{tabularx}{1.0\textwidth}{m{1cm}m{4cm}m{4cm}}
    & \begin{minipage}{.4\textwidth}
    \centering \%error\\
    sinkhorn\_Sp
    \end{minipage}
    &
      \begin{minipage}{.4\textwidth}
      \centering \%error\\ sinkhorn\_D1D2
      \end{minipage}
      \\
        $n\times n$
        &
         \includegraphics[width=.4\textwidth]{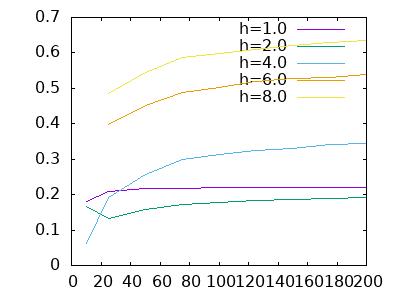}&
         \includegraphics[width=.4\textwidth]{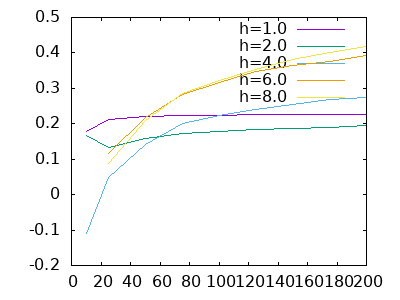}\\
    $n\times 2n$&
    \includegraphics[width=.4\textwidth]{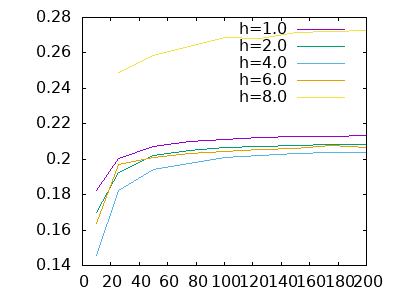}&
    \includegraphics[width=.4\textwidth]{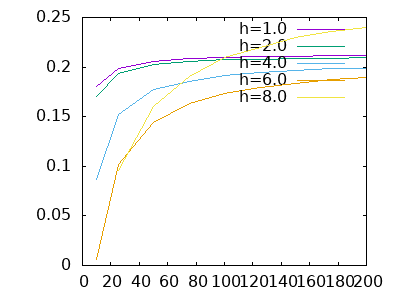}\\
    \end{tabularx}
    
    \caption{Relative errors of sinkhorn\_Sp and sinkhorn\_D1D2 according to the optimal solution. }
    \label{fig:rel_errorh1-8}
\end{figure}

For $h$ greater than $.5$ we observe in Figure~\ref{fig:rel_errorh1-8} that we keep an error of about $20\%$ for $h=1.0$ and $h=2.0$. In these cases the values of the last line and the last column remain comparable with the inner values of the random matrix. However, for larger values of $h$, namely $h\in\{4.0,6.0,8.0\}$ we observe a large increase of the relative error especially in the case of squared matrices. We can conclude from these experiments that our algorithms do not converge to the expected value when the values of the last column/line are very large compared to the inner values. More precisely, when we have:
\[
s_{i,j}\ll s_{n+1,i}+s_{i,m+1} \mbox{ for }(i,j)\in\{1,\dots,n\}\times\{1,\dots,m\}
\]
A simple solution to fix this problem, consists in simplifying the similarity matrix by removing (setting to a low value) any entry $(i,j)$, $i\in\{1,\dots,n\}$ and $j\in\{1,\dots,m\}$ such that $(i,j)$ cannot belong to any optimal solution. Given the similarity matrix $S$, such entries are characterized by $s_{i,j}<s_{n+1,j}+s_{i,m+1}$.  In such cases, the substitution of $i$ onto $j$ may be advantageously replaced by the removal of $i$ and the insertion of $j$. This last point forbids the assignment of $i$ onto $j$ in any optimal $\epsilon$-assignment.

\begin{figure}
    \centering
    \begin{tabularx}{1.0\textwidth}{m{1cm}m{4cm}m{4cm}}
    & \begin{minipage}{.4\textwidth}
    \centering \%error\\
    sinkhorn\_Sp
    \end{minipage}
    &
      \begin{minipage}{.4\textwidth}
      \centering \%error\\ sinkhorn\_D1D2
      \end{minipage}
      \\
        $n\times n$
        &
         \includegraphics[width=.4\textwidth]{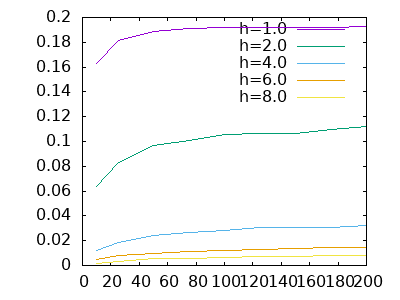}&
         \includegraphics[width=.4\textwidth]{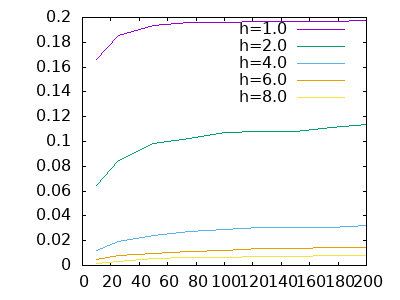}\\
    $n\times 2n$&
    \includegraphics[width=.4\textwidth]{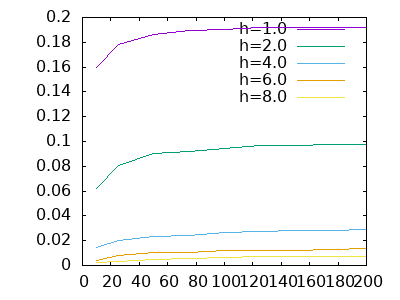}&
    \includegraphics[width=.4\textwidth]{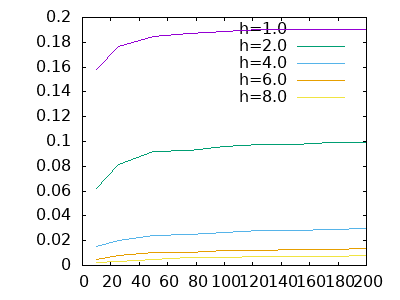}\\
    \end{tabularx}
    
    \caption{Relative errors of sinkhorn\_Sp and sinkhorn\_D1D2 according to the optimal solution after a simplification of the similarity matrix. }
    \label{fig:rel_error_normh1-8}
\end{figure}

Figure~\ref{fig:rel_error_normh1-8} represents the relative errors according to the optimal solution performed by Sinkhorh\_Sp and sinkhorn\_D1D2 using this simplification of the similarity matrix. In this experiment the "low value" replacing any entry of the matrix $S$ which can not be included in any optimal solution has been fixed to $10^{-4}$. One can first observe that we get the same behavior for the squared ($n\times n$) case and the rectangular one ($n\times 2n$). We can further observe that all errors remain below $20\%$ for all sizes. Moreover, the relative error appear to be decreasing as a  function of $h$ for both algorithms. This may be explained by the fact that as $h$ get higher, the simplified matrix becomes more and more trivial. Indeed for largest values of $h$, simplified similarity matrices correspond to trivial matrices with a constant value (equal to $10^{-4}$ in this experiment) for all entries $(i,j)$ in $\{1,\dots,n\}\times\{1,\dots,m\}$ and a last row and column which remains unchanged and greater than $10^{-4}$ by several orders of magnitude. In such cases our algorithms converge immediately to the optimal solution which correspond to the removal of all elements in $\{1,\dots,n\}$ and the insertion of all elements in $\{1,\dots,m\}$.

\begin{figure}
    \centering
    \begin{tabularx}{1.0\textwidth}{m{1cm}m{4cm}m{4cm}}
    & \begin{minipage}{.4\textwidth}
    \centering 
    sinkhorn\_Sp
    \end{minipage}
    &
      \begin{minipage}{.4\textwidth}
      \centering  sinkhorn\_D1D2
      \end{minipage}
      \\
        graphic card
        &
         \includegraphics[width=.35\textwidth]{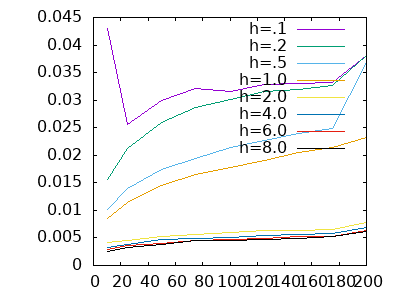}&
         \includegraphics[width=.35\textwidth]{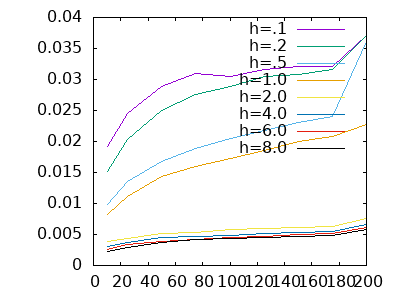}\\
    cpu&
    \includegraphics[width=.35\textwidth]{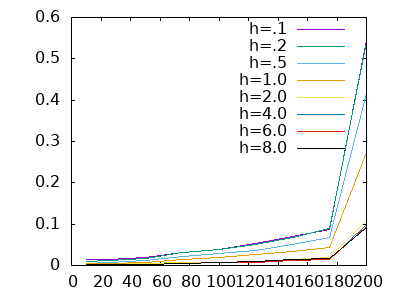}&
    \includegraphics[width=.35\textwidth]{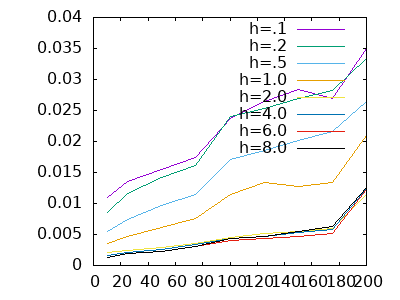}\\
    \end{tabularx}
    
    \caption{Execution times of  sinkhorn\_Sp and sinkhorn\_D1D2 on squared $n\times n$ matrices using both  graphic card and cpu computation. }
    \label{fig:exe_timenxn}
\end{figure}

\begin{figure}
    \centering
    \begin{tabularx}{1.0\textwidth}{m{1cm}m{4cm}m{4cm}}
    & \begin{minipage}{.4\textwidth}
    \centering 
    sinkhorn\_Sp
    \end{minipage}
    &
      \begin{minipage}{.4\textwidth}
      \centering  sinkhorn\_D1D2
      \end{minipage}
      \\
        graphic card
        &
         \includegraphics[width=.35\textwidth]{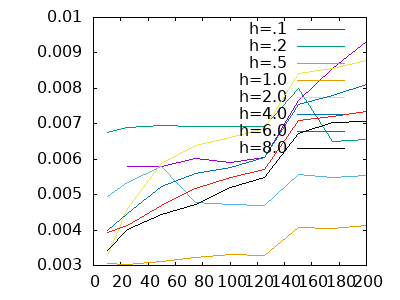}&
         \includegraphics[width=.35\textwidth]{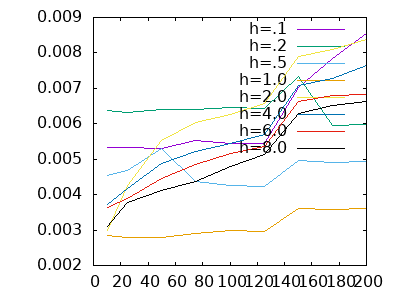}\\
    cpu&
    \includegraphics[width=.35\textwidth]{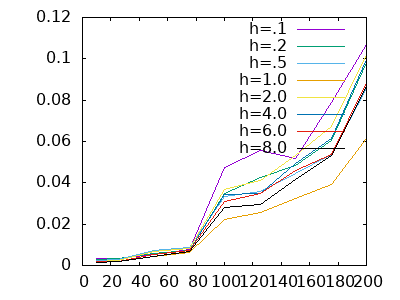}&
    \includegraphics[width=.35\textwidth]{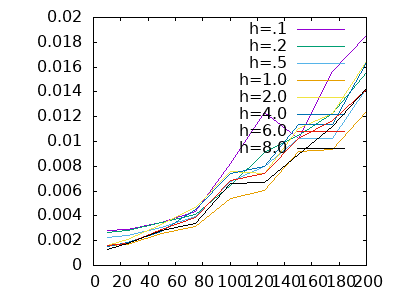}\\
    \end{tabularx}
    
    \caption{Execution times of  sinkhorn\_Sp and sinkhorn\_D1D2 on  $n\times 2n$ rectangular matrices using both  graphic card and cpu computation. }
    \label{fig:exe_timenx2n}
\end{figure}

\subsection{Execution times of our algorithms}

The execution times of both sinkhorn\_Sp and sinkhorn\_D1D2 computed either on a graphic card (Nvidia Quadro P2000) or on a CPU (Intel Core i5 650@3.2GHz) are displayed in  Figures~\ref{fig:exe_timenxn} for $n\times n$ squared matrices  and in Figure~\ref{fig:exe_timenx2n}  for $n\times 2n$ rectangular matrices. In both figures and for each size, both algorithms have been run $100$ times and the execution times have been averaged. 

Considering the squared case (Figure~\ref{fig:exe_timenxn}) we can see that on graphics cards both sinkhorn\_Sp and sinkhorn\_D1D2 take roughly the same amount of time. This is clearly not the case on CPU where the updates of the matrix $S_p$ instead of the two diagonal matrices $D_1$ and $D_2$ induce a large difference, by a factor greater than 10, between the execution times of sinkhorn\_Sp and sinkhorn\_D1D2. We can further observe that if we do not observe major differences between the cuda and cpu execution times for sinkhorn\_D1D2 this is clearly not the case for sinkhorn\_Sp which is more than 10 times accelerated by the use of the graphic card. Finally, we can note that the execution times are decreasing as a function of $h$. As previously, this last point is due to the fact that as $h$ get higher the simplified similarity matrices become more and more trivial and our iterative algorithms need less and less iterations to converge.

Concerning the rectangular case (Figure~\ref{fig:exe_timenx2n}), we observe the same trends than in the squared case. However the ratio between the execution times of sinkhorn\_Sp and  sinkhorn\_D1D2 on CPU is, in this case, about 6. Moreover, we observe for such matrices a factor approximately equal to 2 between the  execution times of   sinkhorn\_D1D2 on  GPU and CPU. As in the squared case the CUDA implementation provides a large speedup for sinkhorn\_Sp. Finally, the execution times of both algorithms decrease as $h$ increase. This phenomenon already encountered in the squared case is due to the same reasons.  

\section{Conclusion}
\label{sec:conclu}

We have presented in this report (Section~\ref{sec:algo}) two algorithms. The proof of their convergence is provided in Section~\ref{sec:construct} while conditions of the existence and uniqueness of the limits are provided in Section~\ref{sec:ExistUniq}. Section~\ref{sec:sim_to_cost} provides simple methods to transform the sum maximization problem addressed by our algorithms into sum minimization (minimization of a sum of costs). 

As shown in Section~\ref{sec:xp}, these algorithms  provide an approximate solution to the Linear Sum Assignment Problem with Edition (LSAPE).  The relative error of these algorithms compared to the optimal solutions is similar, and even much lower in some cases, to the relative error between the classical Sinkhorn and the optimal solutions to the Linear Sum Assignment Problem (LSAP). The main difference between these algorithms and the Hungarian based algorithms providing the optimal solution is that our algorithms are iterative and differentiable and may thus be easily inserted within a backpropagation based learning framework such as artificial neural networks.

Compared to the LSAP, the LSAPE problem allows to manage  assignments between sets of different sizes by allowing the possibility to reject some elements from the matching through the use of insertion/deletion operations. Let us note that these insertion/deletion operations are integrated to the matching algorithm. The LSAPE  thus avoids any artificial preprocessing step consisting in the selection of the more promising objects of both sets for matching. Our algorithms output an $\epsilon$ bi-stochastic matrix which mainly differs from  the results provided by~\cite{peyre2019computational,pmlr-v84-genevay18a} through the  explicit notions of insertion and deletion.

Let us finally note that we have required in Section~\ref{sec:algo} that all the coefficients of the last line and column of the similarity matrix should be positive. This requirement is certainly too strong and weaker conditions may certainly  be established by future works. An alternative future research direction, consists in addressing  the LSAPE problem using the recent advances in Graph Neural Networks (GNN). 
\bibliographystyle{plain} \bibliography{biblio}

\end{document}